\documentclass[pdflatex,sn-mathphys]{sn-jnl}

\jyear{2021}%

\theoremstyle{thmstyleone}%
\theoremstyle{thmstyletwo}%

\theoremstyle{thmstylethree}%

\usepackage{amssymb}
\usepackage{pifont}
\usepackage{subcaption}
\usepackage{amsmath}

\usepackage{listings}
\usepackage{xcolor}
\definecolor{codegreen}{rgb}{0,0.6,0}
\definecolor{codegray}{rgb}{0.5,0.5,0.5}
\definecolor{codepurple}{rgb}{0.58,0,0.82}
\definecolor{backcolour}{rgb}{0.95,0.95,0.92}

\lstdefinestyle{mystyle}{
    backgroundcolor=\color{backcolour},   
    commentstyle=\color{codegreen},
    keywordstyle=\color{magenta},
    numberstyle=\tiny\color{codegray},
    stringstyle=\color{codepurple},
    basicstyle=\ttfamily\footnotesize,
    breakatwhitespace=false,         
    breaklines=true,                 
    captionpos=b,                    
    keepspaces=true,                 
    numbers=left,                    
    numbersep=5pt,                  
    showspaces=false,                
    showstringspaces=false,
    showtabs=false,                  
    tabsize=2
}
\begin{document}


\title[AugLiChem]{AugLiChem: Data Augmentation Library of Chemical Structures for Machine Learning}


\author[1]{\fnm{Rishikesh} \sur{Magar}}
\equalcont{These authors contributed equally to this work.}

\author[1]{\fnm{Yuyang} \sur{Wang}}
\equalcont{These authors contributed equally to this work.}

\author[1]{\fnm{Cooper} \sur{Lorsung}}
\equalcont{These authors contributed equally to this work.}

\author[2]{\fnm{Chen} \sur{Liang}}

\author[1]{\fnm{Hariharan} \sur{Ramasubramanian}}

\author[2]{\fnm{Peiyuan} \sur{Li}}

\author*[1,2]{\fnm{Amir} \sur{Barati Farimani}}\email{barati@cmu.edu}

\affil[1]{\orgdiv{Department of Mechanical Engineering}, \orgname{Carnegie Mellon University}, \orgaddress{\city{Pittsburgh}, \state{PA} \postcode{15213}, \country{USA}}}

\affil[2]{\orgdiv{Department of Chemical Engineering}, \orgname{Carnegie Mellon University}, \orgaddress{\city{Pittsburgh}, \state{PA} \postcode{15213}, \country{USA}}}




\abstract{Machine learning (ML) has demonstrated the promise for accurate and efficient property prediction of molecules and crystalline materials. To develop highly accurate ML models for chemical structure property prediction, datasets with sufficient samples are required. However, obtaining clean and sufficient data of chemical properties can be expensive and time-consuming, which greatly limits the performance of ML models. Inspired by the success of data augmentations in computer vision and natural language processing, we developed AugLiChem: the data augmentation library for chemical structures. Augmentation methods for both crystalline systems and molecules are introduced, which can be utilized for fingerprint-based ML models and Graph Neural Networks (GNNs). We  show that using our augmentation strategies significantly improves the performance of ML models, especially when using GNNs. In addition, the augmentations that we developed can be used as a direct plug-in module during training and have demonstrated the effectiveness when implemented with different GNN models through the AugliChem library. The Python-based package for our implementation of Auglichem: Data augmentation library for chemical structures, is publicly available at: \url{https://github.com/BaratiLab/AugLiChem}.}




\maketitle


\section{Introduction}\label{sec1}


Machine learning (ML) models, especially deep neural networks (DNNs) \cite{lecun2015deep}, capable of learning representative features from data, have gathered growing attention in computational chemistry and material science \cite{compchem}. Moreover, advances from areas, such as computer vision (CV) and natural language processing (NLP), have often had synergistic effect on ML research in computational chemistry. This has lead to development of graph neural networks (GNNs) based ML models capable for accurate and efficient crystalline and molecular property prediction \cite{fung2021benchmarking,wu2018moleculenet,MolPeter2019} and fingerprint (FP) based ML models that take chemical FPs \cite{PhysRevB.87.184115_SOAP,huo2017unified_MBTR,lam2017machine_OFM,botu2017machine,rogers2010extended} as input and predict target properties. However, development of the FPs requires manual design and is sometimes time-consuming to calculate, which limits the performance of ML models leading to increased popularity of GNN based models. 
Recently, GNNs \cite{kipf2017semi, xu2018how} have been a prevalence in computational chemistry, as GNNs are capable of learning representations from non-Euclidean graphs of chemical structures directly without manually engineered descriptors \cite{schutt2018schnet,klicpera_dimenet_2020}. GNNs developed for crystalline systems and molecules to predict various target properties and have shown advantage over other ML methods \cite{justin2017neural,yang2019analyzing, wang2021molclr, xie2018cgcnn, karamad2020orbital} and a remarkably high performance. However, ML models, including GNNs and  FP-based models, require large amounts of data for training \cite{hestness2017deep}. Additionally, the availability of large and clean data is an important pre-requisite as the performance of ML models scales with the magnitude of data \cite{zhu2016we}. However, generating large scale properties datasets for crystalline systems or molecules usually requires sophisticated and time-consuming computational or lab experiments \cite{brown2019guacamol}. The experimental data can also be noisy and hard to utilize directly. Thus, the lack of data of has often prevented the use of deep learning models as they suffer from lower reliability in the small data regime \cite{bengio_limit_deep}. 

The issue of data availability also plagues the areas of computer vision (CV) and natural language processing(NLP). To address the issue of data availability, data augmentation techniques have been introduced in CV and NLP, which generates new instances from available training data via some transformations to increase the amount and variance of the data \cite{bitton2021augly}. Such techniques have been investigated for multiple domains, like images and text, and has demonstrated the effectiveness in improving the generalization, performance, and robustness of ML models \cite{shorten2019survey,kobayashi2018contextual, coulombe2018text, pmlr-v119-chen20j}. For images in CV, the fundamental augmentation transformations include cropping, rotation, color distortion, etc. Following the transformations, some works manually design transformations \cite{devries2017improved} or learn the desired mixture of image transformations \cite{zhang2017mixup, cubuk2018autoaugment, yin2019fourier} during training. Augmentations have also been investigated on the representation domain instead of directly on input, including adding noise, interpolating, and extrapolating
of image representations \cite{devries2017dataset, konno2018icing}. In addition, data augmentations have been widely used for texts in NLP \cite{feng2021survey}, like random deletion, insertion, and swap. One common method is to replace words with their synonyms \cite{zhang2015character} or relative word embeddings \cite{wang2015s, fadaee2017data}. Other augmentations focus on manipulation of the whole sentence, such as back translation \cite{sennrich2015improving} and generative models \cite{kafle2017data}. Recent works have also found that simple augmentations (e.g., cropping, resizing, color distortion for images and token masking for texts) significantly benefit representation learning through self-supervised pre-training \cite{devlin2018bert, chen2020simple, tian2020makes, gao2021simcse}. Motivated by the success of data augmentation on CV and NLP, some works in the area of molecular machine learning investigate augmentation for string-based molecular descriptors (e.g., SMILES \cite{weininger1988smiles}) \cite{bjerrum2017smiles,lambard2020smiles} or conformational oversampling \cite{hemmerich2020cover}. However, data augmentation is still under-explored for chemical structures, especially for graph-based ML models. Unlike images or text, crystalline and molecular systems form graphical structures and follow chemical rules, like composition of motifs and boundary conditions. Therefore, direct leverage of existing augmentations, such as image resizing and cropping, does not work for chemical structures. This leads us to the question: can we develop data augmentation techniques for crystalline systems and molecules to boost the performance of ML models, especially GNNs? 




\begin{figure}[htb!]

\centering
    \includegraphics[width=1\textwidth, keepaspectratio=true]{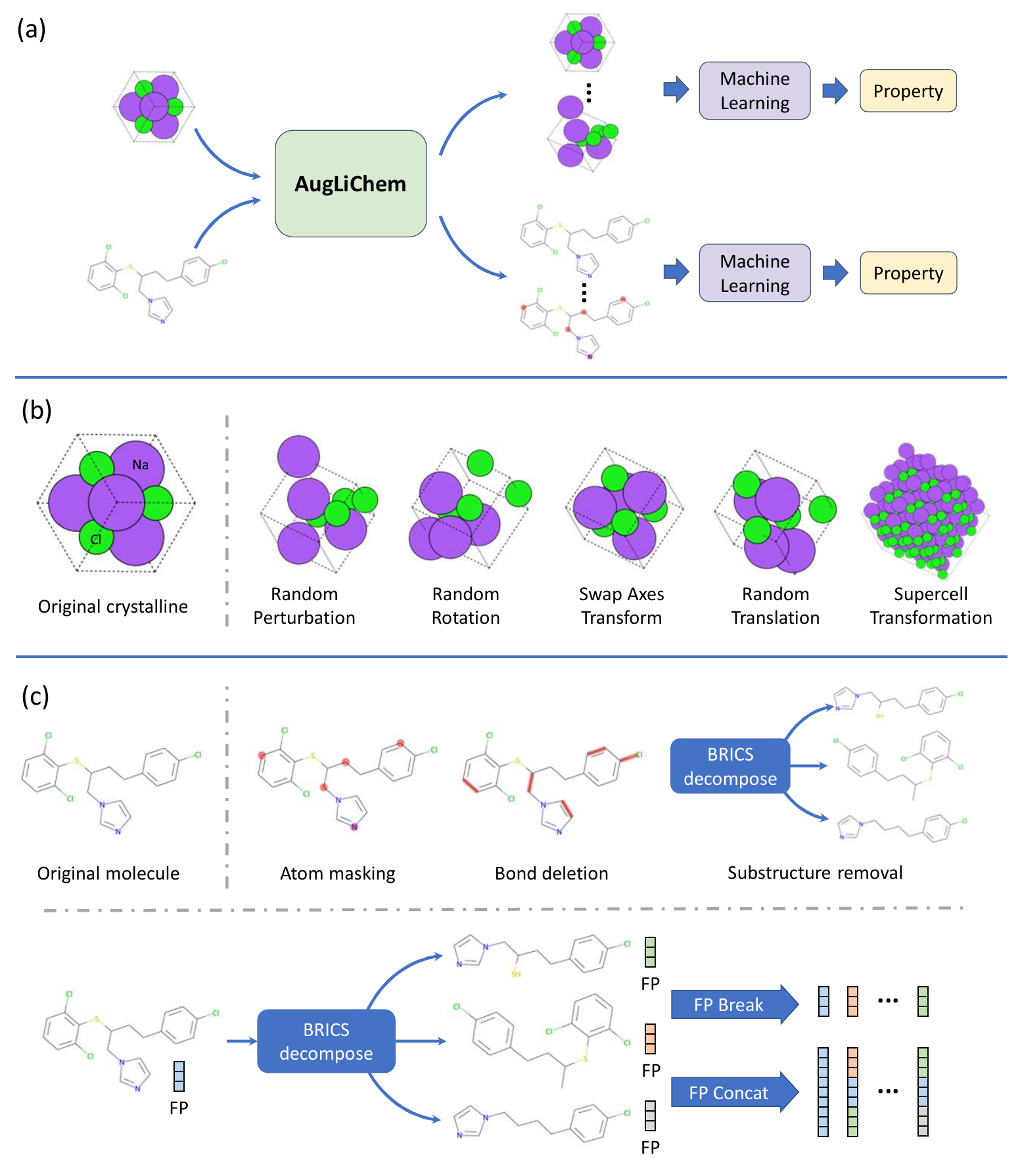}


    
    
    

\caption{Overview of the augmentations for chemical structures. (a) Framework of AugLiChem package. (b) Augmentations of crystalline materials for both conventional ML models and GNNs. (c) Augmentations of molecules for GNNs (top row) and augmentations for fingerprint-based ML models (bottom row).}
\label{fig:overview}
\end{figure}

We develop AugLiChem: data augmentation library for chemical structures, which boost the data availability of crystals and molecules and helps improve the performance of ML models. The augmentations can be implemented as simple plug-ins to FP-based ML models as well as GNNs. The whole pipeline of AugLiChem is illustrated in Figure~\ref{fig:overview}(a), which first takes in the chemical structures, either crystalline or molecular, and creates augmented data instances. The augmented data is then assigned with the same label as the original instance. The original chemical instances, together with the augmented ones, are fed into ML models for property prediction. For crystals, five augmentation methods are proposed: random perturbation, random rotation, swap axes, random translation, and supercell transformation, which can be direct plug-ins for both conventional ML models and graph-based GNNs as shown in Figure~\ref{fig:overview}(b). Similarly for molecules, we introduce three molecule graph augmentations: atom masking, bond deletion, and substructure removal. Since fingerprint (FP) calculation are different between molecules and crystalline materials, molecule graph augmentations cannot be directly utilized for FP-based ML models. We consider two extra augmentation methods for molecule FPs. Figure~\ref{fig:overview}(c) illustrates both the graph and FP augmentations for molecules. Experiments on various benchmarks have demonstrated the effectiveness of our data augmentations for both crystals and molecules. Especially for GNNs, augmentation significantly boosts the performance of chemical property prediction tasks. Finally, for convenient use of our data augmentation techniques, we develop an open-source library for chemical structure augmentation based on Python as part of this work. Both crystal and molecule augmentations can be implemented with a few lines of code. Different GNN models and widely-used benchmarks are also included in the package. The package is largely self contained with tutorials describing the functionality of different package features and example codes to run them.

\section{Methods}\label{sec4}

\subsection{Crystalline Augmentations}

We treat a crystalline system as an undirected graph with atoms as the nodes and the bonds as the edges. The crystal graph can be denoted by $G_{cr}=(V_{cr},E_{cr})$, where $V_{cr}$ are the nodes of the graph denoting the atom sites in a crystal structure. The idea of a bond in crystalline systems is not as concretely defined as in molecular graphs. Therefore, to construct the edges between the nodes, all atoms within a radius are considered to be bonded. The strength of the bond is then determined by the amount of distance from the central atom and edge features are encoded accordingly. After constructing the graph, we treat as a input feature to the GNN model, the GNN model then learns from the graph and predicts the material property. 

We implement 5 different types of crystal augmentations as a part of this work. The augmentations we implement include random perturbation, random rotation, swap axes, random translate, supercell transformation. Each of the augmentation that we implement creates a different view of the crystal enabling the machine learning models particularly GNNs to take advantage of the large data set and generate accurate representations the crystal. We show that using augmentations we can improve the prediction performance of GNNs on all the data sets and also improve performance of the shallow machine learning models for a majority of the data sets. We would like to note that these augmentation strategies have been applied only for the training set and not for the validation and test sets.

\subsubsection{Random Perturbation}
In the random perturbation augmentation all the sites in the crystalline system are randomly perturbed by a distance between 0 to 0.5 \AA. This augmentation is especially useful in breaking the symmetries that exist between the sites in the crystals.
\subsubsection{Random rotation}
In the random rotation transform, we randomly rotate the sites in the crystal between 0 to 360 degrees. To generate the augmentation we initially use the random perturbation augmentation to generate the initial structure which is then rotated randomly.
\subsubsection{Swap Axes}
In the swap axes augmentation strategy, we swap the coordinates of the sites in the crystal between two axes. For example, we may swap the x and the y axes coordinates or the y and z axes coordinates. The swap axes transform greatly displaces the locations of all the sites in the crystal. This augmentation strategy has been inspired from work done by Kim et al.\cite{kim2020generative_gan}

\subsubsection{Random Translate}
The random translate transform randomly displaces a site in the crystal by a distance between 0 to 0.5 \AA. In this work, we randomly select 25\% of the sites in the crystal and displace them. This creates an augmentation different from the random perturb augmentation as not all the sites in the crystal are displaced.

\subsubsection{Supercell Transformation}
The supercell transformation produces a supercell of the crystalline system. The distinct feature of the supercell of the crystal is that after transformation the supercell represents the same crystal with a larger volume. There exists a linear mapping between the basis vectors of crystal and the basis vectors of the supercell. 


\subsection{Molecule Augmentations}

For molecules, we introduce augmentations that can be applied when using FP based models and GNNs. A molecule can be embedded to a unique FP, like ECFP \cite{rogers2010extended} and RDKFP. We introduce two molecular fingerprint (FP) augmentation techniques, FP Break and FP Concatenation (FP Concat for brevity), are proposed for FP-based ML models.

In addition, for GNN models, a molecular graph $G_m=(V_m,E_m)$ is defined where nodes $V_m$ denote atoms and edges $E_m$ denote chemical bonds between atom pairs \cite{xiong2019pushing,Hu2020Strategies}. For $v_i \in V_m$, its node feature consists of two components, one-hot embedding of the atomic type $a_i$ and the chirality $c_i$, namely $v_i = [a_i, c_i]$. For $e_{ij} \in E_m$, its edge feature also consists of two components, one-hot embedding of the bond type $b_{ij}^t$ and one-hot embedding of the bond directional $b_{ij}^d$, namely $e_{ij}=[b_{ij}^t, b_{ij}^d]$. Both node and edge features can be enlarged with more attributes, like valency and aromaticity. 
Three molecular graph augmentation techniques, node masking, bond deletion, and substructure removal, are also introduced and can be direct plugins to various GNN models \cite{you2020graphcontrstive,wang2021molclr}. 

\subsubsection{Fingerprint Break}

A molecule is broken into multiple-level fragments through Breaking Retrosynthetically Interesting Chemical Substructures (BRICS) \cite{degen2008art} decomposition, where each fragment reserves one or more functional groups from the original chemical structure. Notice that BRICS decomposition follows a tree-structure such that high-level fragments can contain low-level fragments. Fragments obtained from BRICS along with the molecule are featurized through molecular fingerprints, like RDK topological fingerprint (RDKFP) \cite{greg2006rdkit} and extended-connectivity fingerprint (ECFP) \cite{rogers2010extended}. We calculate the Tanimoto similarity score \cite{bajusz2015tanimoto} between fragments and the molecule and only keep the fragments with score greater than a threshold $S$. This means fragments which contain important functional groups and share similar topology with the original molecule are retained. Since the property of a molecule is heavily determined by the functional groups, fragments and the molecule are assumed to share similar properties \cite{mcnaught1997compendium,smith2020march}. In FP Break, each fragment is assigned with the same label as the molecule in the training set. In the validation and test set, only the original molecules are used. 

\subsubsection{Fingerprint Concatenation}

On the other hand, FP Concat follows the same BRICS decomposition as the FP Break, except that the former concatenates randomly picked $K$ FPs from fragments and the molecule, and assigns the same label as the original molecule to each concatenated FP. Additionally, FP Concat includes a replicated FP that repeats the FP of the original molecule $K$-times, so that the original FP can be used with the trained models. In the testing phase, models are tested only on the replicated FPs from the validation and test sets. In our implementation of both FP Break and FP Concat, $S$ is set as 0.6 and $K$ as 4. In atom masking, a node $v_i$ is randomly masked by replacing the node feature $a_i$ with the mask vector $m$. Bond deletion removes an edge $e_{ij}$ from the molecular graph with a certain probability. And substructure removal follows the BRICS decomposition in FP augmentations, where molecular graphs are created based on the decomposed fragments from BRICS and are assigned with the same label as the original molecule. 

\subsubsection{Atom Masking}

In atom masking, a node $v_i$ is randomly masked by replacing the node feature $a_i$ with the mask vector $m$, which is a unique vector different from all the node features in the database. Thus, the GNNs lose the atomic information concerning the masked atoms during training, which forces the model to learn the correlation between atoms through messages passing from the adjacencies. 

\subsubsection{Bond Deletion}

Bond deletion randomly removes an edge $e_{ij}$ from the molecular graph given a certain probability, so that the node pair $(v_i,v_j)$ are not directly connected within the molecular graph. This means the model is forced to aggregate information from atoms that are not adjacent in the molecular graph. Bond deletion relates to the breaking of chemical bonds which prompts the model to learn correlations between the involvements of one molecule in various reactions. 

\subsubsection{Substructure Removal}

Substructure removal follows the BRICS decomposition in FP augmentations, where molecular graphs are created based on the decomposed fragments from BRICS and are assigned with the same label as the original molecule. Fragment graphs contain the one or more important functional groups from the molecule, and GNNs trained on such augmented data learns to correlate target properties with functional groups.

\subsection{Graph Neural Networks for Chemical Structures}
\begin{figure}
\centering
    \begin{subfigure}{0.95\textwidth}
      \centering
      \includegraphics[width=\linewidth]{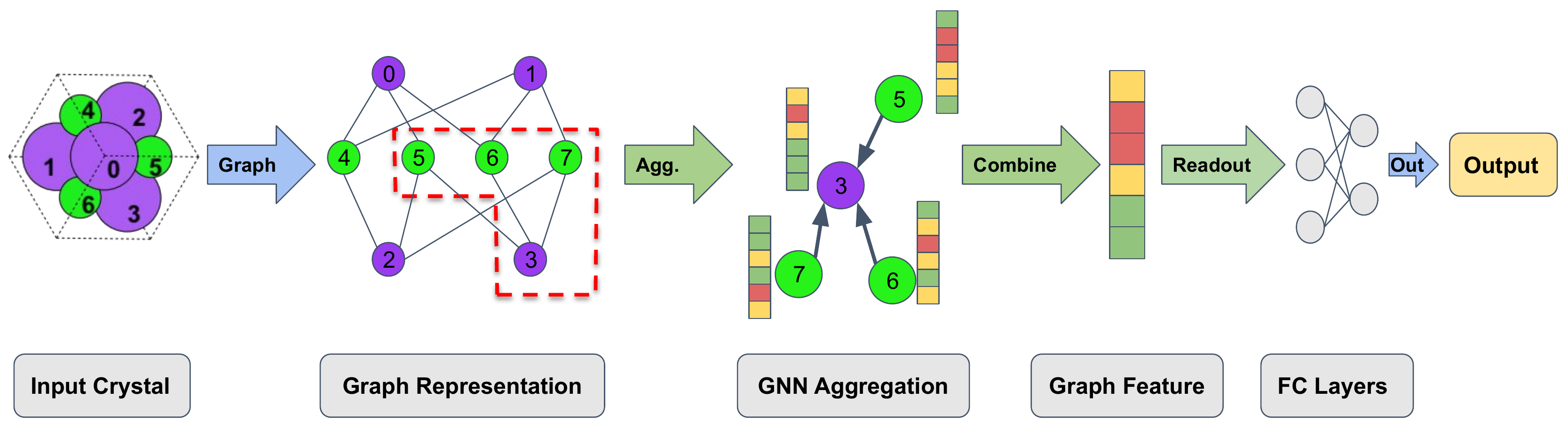}
      \caption{}
    \end{subfigure}
    \begin{subfigure}{0.95\textwidth}
      \centering
      \includegraphics[width=\linewidth]{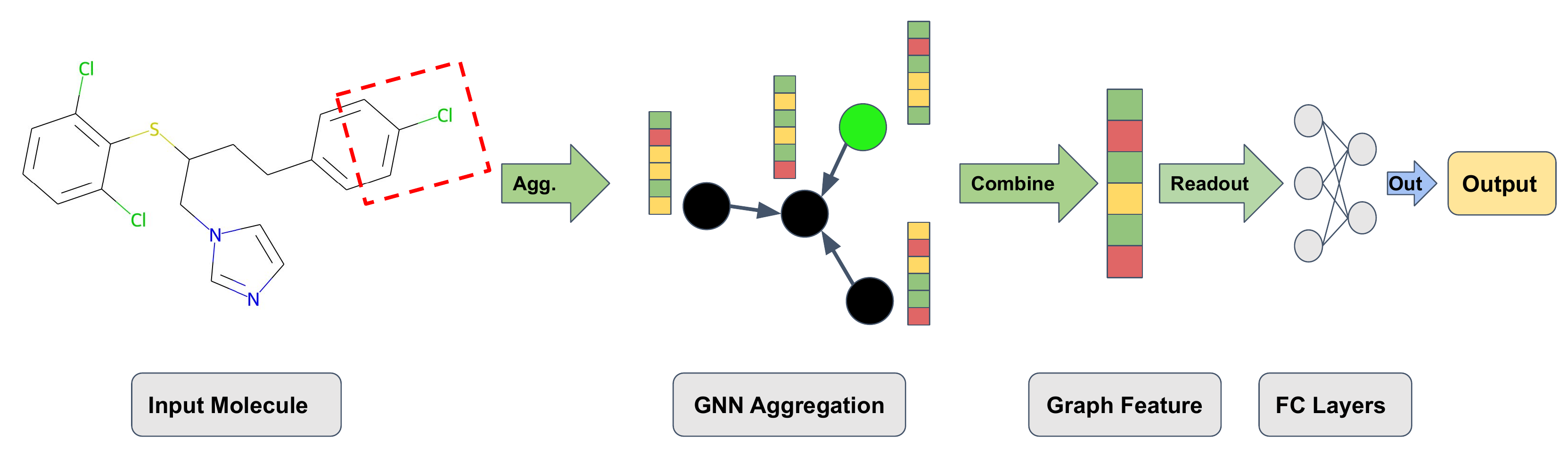}
      \caption{}
    \end{subfigure}
\caption{GNNs for crystalline systems and molecules. (a) In the crystal ML pipeline, we start with an input crystal. The graph representation is constructed with atoms representing the nodes $V$ and the bonds representing the edges $E$ connecting adjacent nodes. Once the graph is constructed, the graph convoluational operations (shown in green) are conducted followed by the fully-connected layers to predict the target property. (b) Similarly to the crystalline graph, a molecule graph contains nodes representing atoms, while edges are defined as chemical bonds between atoms. The built molecule graph is then fed to the GNNs and fully-connected layers to predict molecular properties.}
\label{fig:numbered_nacl}
\end{figure}

Modern GNNs updates node-level representations through the aggregated message passing between nodes and extract graph-level representations through readout \cite{wu2020comprehensive}.
Recently, GNNs have been widely investigated in modeling chemical structures \cite{zhou2020graph}.
GNNs have been succesfully used on the molecular graphs to learn FPs \cite{duvenaud2015conv,kearnes2016molecular}, predict molecular properties \cite{justin2017neural,yang2019analyzing,huang2020skipgnn, wang2021molclr}, and generate target molecules \cite{de2018molgan,you2018graph,imrie2020deep}. Chemical structures, both organic molecules and inorganic structures, are represented as graphs. In this work, each node represents an atom \cite{Hu2020Strategies} and edges can be defined as covalent bonds (molecules) \cite{xiong2019pushing,wang2021molclr,duvenaud2015conv} or connection to adjacent atoms (crystals) \cite{xie2018cgcnn,klicpera_dimenet_2020}. This makes molecular graphs naturally translationally and rotationally invariant.
Examples of building a graph from a crystalline system and a molecule is illustrated in Figure~\ref{fig:numbered_nacl}. 


GNNs utilize a neighborhood aggregation operation, which update the node representation iteratively. The update of each node in a GNN layer is given in Equation~\ref{eq:aggregate}:
\begin{align}
\begin{split}
    \pmb{a}_v^{(k)} &= \text{AGGREGATE}^{(k)}(\{\pmb h_u^{(k-1)}: u \in \mathcal{N}(v)\}), \\
    \pmb h_v^{(k)} &= \text{COMBINE}^{(k)}(\pmb h_v^{(k-1)}, \pmb a_v^{(k)}),
\end{split}
\label{eq:aggregate}
\end{align}
where $\pmb h_v^{(k)}$ is the feature of node $v$ at the $k$-th layer and $\pmb h_v^{(0)}$ is initialized by node feature $\pmb x_v$. $\mathcal{N}(v)$ denotes the set of all the neighbors of node $v$. GNNs implemented in this work have multiple aggregation and combination functions, including mean or max pooling \cite{kipf2017semi,hamilton2017inductive}, fully-connected (FC) layers \cite{xu2018how}, recurrent neural network (RNN) \cite{hamilton2017inductive,li2019deepgcns}, Gaussian weighed summation \cite{xie2018cgcnn, karamad2020orbital}, and attention mechanism \cite{velickovic2018graph, xiong2019pushing}.
Other works exploits edge features in the message passing to leverage more information \cite{gong2019exploiting,Hu2020Strategies}.
To extract a graph-level feature $\pmb h_G$ from the node features, readout operation integrates all the node features among the graph $G$ as given in Equation~\ref{eq:readout}:
\begin{equation}
    \pmb h_G = \text{READOUT}(\{\pmb h_u^{(k)}: v \in G\}).
    \label{eq:readout}
\end{equation}
Common readout operations can be the summation, averaging, or max pooling over all the nodes \cite{xu2018how}. More recently, differentiable pooling readout has been devloped\cite{ying2018diffpool,gao2019graph}.
Given $\pmb h_G$, either FC layers \cite{Hu2020Strategies} or RNNs \cite{xiong2019pushing} are developed to map the graph feature to the predicted property. 

Unlike two-dimensional (2D) molecular graphs, crystal graphs usually require the 3D positional data as the input. Crystal graphs treat the atoms in the crystal as nodes and edges in crystal graphs are defined as relationship between neighboring atoms\cite{schmidt2019recent,schleder2019dft}. Following the setting, various aggregation operations \cite{xie2018cgcnn,chen2019graph} and crystalline features \cite{st2019message,karamad2020orbital,park2020developing} have been investigated in predicting crystal properties using GNNs. 
Additionally, some works also consider 3D positional features in molecular GNNs for property predictions concerning quantum mechanics \cite{schutt2018schnet, klicpera_dimenet_2020}.

In this work, we use data augmentation strategies for multiple GNN backones for both the crystalline and molecular systems. For the crystalline systems we use the CGCNN \cite{xie2018cgcnn}, MEGNET\cite{chen2019graph}, SchNet\cite{schutt2017schnet}, GINE models\cite{Hu2020Strategies}. The CGCNN model builds a crystal graph by taking the atoms as the nodes and the modelling the interaction between the atoms as the edges. To describe the edge features the CGCNN model uses the  Gaussian featurization. The model uses 3 GCN layers to aggregate node and edge features and does the material property prediction. The MEGNET model takes as input the atom features, bond features and the current state, the state is representative of the global feature vector, in the update the bond information is updated first followed by atom feature vector and global state vector. The update strategy is performed in the MEGNET blocks introduced in the architecture and fully connected layers are subsequently used to perform material property prediction. The SchNet model uses the continuous convolutional filter layers to model the molecular structures. The SchNet model uses a feature map to represent the atom feature and takes into account the atomic positions of the atoms and their nuclear charges. The model uses a a combination of interaction block and the atom-wise layers to generate the representations that can be used for prediction tasks. Finally, the GINE model utilizes an MLP to perform the weighted aggregation of node and edge features. We use mean pooling as the readout function for the GIN and use the pooled vector extract the representations of the crystalline system that can be used used for property prediction. 

For molecules, we implement GCN \cite{kipf2017semi}, GIN \cite{xu2018how, Hu2020Strategies}, DeeperGCN \cite{li2019deepgcns, li2020deepergcn}, and AttentiveFP \cite{xiong2019pushing} as GNN backbones to predict molecular properties. These models follow the message passing framework but differ in detailed aggregation and combination operations. GCN aggregates all the neighbors through mean pooling followed by a linear mapping with ReLU activation. GIN, on the other hand, sums up all the neighboring node features and develops an MLP afterwards. However, such GNN models suffer from the vanishing gradient problem when number of graph convolutional layers increases \cite{li2019deepgcns, li2020deepergcn}. To this end, DeeperGCN introduces skip connections between layers. AttentiveFP utilizes the attention mechanism and RNNs to aggregate neighbor information and update node features, which reaches SOTA on several challenging molecular property benchmarks. These models cover a wide variety of GNN frameworks and greatly demonstrate the effectiveness of our molecule augmentation techniques. 

\subsection{Training Details}


We benchmark our augmentation methods on different crystalline and molecule datasets. For the five crystalline benchmarks, instances are first randomly split into train set and test set. The ratio of train set and test set is  4:1. We further split the training data into training and validation set in the ratio of 4:1. Finally, we have 64\% of the data as training data, 16\% of the data as validation data and the remaining 20\% as the test set. The percentages of train/validation/test sets have been kept the same baseline models which are trained without augmented data and the models trained with augmented data. Only the data in the training set has been augmented, we do not augment the validation and test set data.

For the twelve molecule benchmarks, instances are split into train/validation/test by the ratio of 8:1:1 via scaffold splitting, which provides a challenging yet realistic setting \cite{Hu2020Strategies, wang2021molclr}. Augmentations are implemented on the training sets only while validation and test sets stay intact. Each benchmark is split into the same training/validation/test sets, and same augmentations are implemented on instances on the training set if applied. So that all the ML models are trained and test on the same instances for fair comparison.  
For example, for Band Gap benchmark containing 26,709 crystalline instances, it is split into train/validation/test sets and each includes 17,093/4,274/5,342 instances. If five augmented instances are created for each crystalline training data, this gives 85,465 augmented instances in addition to 17,093 instances in the original training set, which adds up to total 102,558 instances in the augmented training set. 
Trained ML models are test on validation sets to select the best performing model and results on the test set are reported in this work.

\section{Results and Discussions}\label{sec2}


\subsection{Investigation of Crystalline Augmentations}\label{subsec2}




To improve the performance of ML models, especially GNNs, we introduce five data augmentation strategies that enhance the amount of data available for training, namely random perturbation, rotation, random swap axes, random translation and supercell transformation. 
To evaluate the effects of using these augmentations when training GNN models, we benchmark the performance of the different models with data augmentation on five different datasets and observe performance gains in all the cases. The five datasets that we include in our study are: Band Gap\cite{doi:10.1063/1.4812323_MP}, Fermi Energy \cite{doi:10.1063/1.4812323_MP}, Formation Energy \cite{doi:10.1063/1.4812323_MP}, Lanthanides\cite{lam2017machine_OFM}, Perovskites\cite{C1EE02717D_castelli}. We implemented 3-fold cross validation strategy to have a better estimate of the model performance, the MAE and the standard deviation of our experiments are demonstrated in \ref{tb:crystal_regression}. The models trained on the augmented data are denoted with subscript "aug". For example, the CGCNN model trained on the augmented data is denoted by CGCNN\textsubscript{aug}. We observe performance gains on all the data sets for all the models validating the usefulness of data set augmentations to improve the performance of the models. We observe a gain between 15\% - 33\% for the five data sets with CGCNN. Similarly, a performance gain between 6\% - 46\% is observed for the MEGNET model. For the SchNet model the improvements lie between 6\%  - 40\% and GIN model shows improvement between 2\% to - 36\%. These improvements indicate the gain in performance of models when trained with augmentations. Additional information on datasets and training details for the GNN model are available in supplementary information (Section - D and Section - F ). We also use the augmentations with shallow machine learning algorithms and examine the performance of these models on the five data sets. To featurize the crystalline systems, we use the AGNI Fingerprints from Botu et al\cite{botu2017machine}. The AGNI fingerprints is a vector of size 32 that represents the crystalline systems. The vector can then be used as feature for the conventional machine learning algorithms like Random Forest and SVM. The results obtained for SVM and RF, however, do not show improvement for all the data sets as in the case of GNN algorithms. We can conclude empirically the effects of data augmentation are more pronounced for for deep learning based architectures than the conventional machine learning algorithms.
\begin{table}[h]
\begin{center}
\tiny
\caption{Results of different GNN models on benchmark datasets with and without crystalline systems data augmentation. Only the data points in the training set have been augmented. Both the average and standard deviation of the MAEs are reported. }
\label{tb:crystal_regression}
\begin{tabular}{@{}l|lllll@{}}
\toprule
\multicolumn{1}{l|}{Dataset} & Band Gap & {Fermi Energy} & Formation Energy & Lanthanides & Perovskites \\

\multicolumn{1}{l|}{\# of data points} & 26709 & 27779 & 26078 & 4133 & 18928  \\

\multicolumn{1}{l|}{\# of Aug data points} & 68372 & 71112 & 66756 & 10580 & 48452  \\
\cline{1-6}
Models & \multicolumn{5}{c}{Prediction on test set (Random Split)} \\
\midrule
RF & 0.9276(0.006)  & 1.128(0.012)  & 0.4474(0.008) & 0.4676(0.01) & 0.4858(0.004)\\
RF\textsubscript{aug} & 0.9382(0.009) & 1.1417 (0.011) & 0.4516(0.006) & 0.4603(0.01) & 0.4823 (0.004) \\

SVM & 1.0520(0.008)  & 1.2484(0.012) & 0.5167(0.006)& 0.8885(0.021) & 0.5280(0.005)\\
SVM\textsubscript{aug} & 1.0474(0.008) & 1.2674 (0.012)& 0.5225 (0.006) & 0.8688(0.02) & 0.5216(0.004)  \\
CGCNN &  0.4588(0.007) & 0.5054(0.010) & 0.0565(0.001) & 0.109(0.002) & 0.0743(0.0005)\\
CGCNN\textsubscript{aug} & \textbf 0.3692(0.001) & 0.4271(0.006) & 0.0380(0.001) & 0.0750(0.002) & 0.0496(0.001) \\
GIN & 0.6006(0.038) & 0.7978(0.014) & 0.1090(0.007) & 0.1969(0.038) & 0.3800(0.008) \\
GIN\textsubscript{aug} &0.5314(0.046) & 0.6054(0.15)  & 0.0771(0.003) & 0.1243(0.003) & 0.3719(0.002)\\
MegNet & 0.4774(0.017) & 0.4872(0.010) & 0.0672(0.002) & 0.1859(0.038) & 0.1162(0.004)\\
MegNet\textsubscript{aug} & 0.3682 (0.009)& 0.4542(0.021) & 0.0410(0.001) & 0.1041(0.009) & 0.0627(0.007) \\
SchNet& 0.7769(0.003) & 0.8257(0.021) & 0.3332(0.020) & 0.5644(0.026) & 0.0733(0.003) \\
SchNet\textsubscript{aug} & 0.6672(0.018) & 0.7717(0.026) &0.1986 (0.019)& 0.3706 (0.026) & 0.0521(0.0003) \\
\botrule
\end{tabular}
\end{center}
\end{table}
\subsection{Investigation of Molecule Augmentations}

We consider two molecular fingerprint (FP) augmentations for FP-based ML models and three molecular graph augmentations \cite{you2020graphcontrstive,wang2021molclr} for GNNs. 
As shown in Table~\ref{tb:mol_classification}, Random Forest (RF) and Support Vector Machine (SVM) are implemented on top of RDKFP \cite{greg2006rdkit} and ECFP \cite{rogers2010extended} for 7 classification benchmarks from MoleculeNet \cite{wu2018moleculenet}. All the benchmarks are split through scaffold splitting to provide a more challenging yet realistic way of splitting \cite{Hu2020Strategies}. Within each benchmark, a model is trained for 3 individual runs and both the mean and standard deviation of compute areas under the receiver operating characteristic curve (ROC-AUC) on test sets are reported. All the experimental results for molecules are reported following this setting unless further mentioning. Models with an "aug" subscript indicate models are trained with an augmented data set. In almost all the classification benchmarks, both RF and SVM trained with augmented data sets outperform baseline models trained only on original data sets. For instance, FP augmentations improve the ROC-AUC of SVM by more 6\% on ClinTox and by 3\% on BACE. It is demonstrated that assigning the same label for FP augmented data enlarges the available training set and sharpens the decision boundary of ML models. Also, performances of FP augmentations on 5 regression benchmarks: FreeSolv, ESOL, Lipo, QM7, QM8, are investigated as shown in Table~\ref{tb:mol_regression}. Similar with classifications, all benchmarks are scaffold split. The mean and standard deviation of rooted mean square errors (RMSEs) in three individual runs are reported. In 4 out of 5 regression benchmarks, augmentations help boost the performance of SVM models. However, RFs trained with augmented data fail to improve the RMSE in comparison to baseline results except on ESOL benchmark. This could because unlike classification tasks where labels are usually abstract and sparse, labels for regression are more sensitive and shallow ML models trained on augmented data can be distracted from the ground truth labels. More detailed results of RF and SVM can be found in Supplementary Information (Table S1 and Table S2). 

\begin{table}[h]
\setlength{\tabcolsep}{3pt}
\begin{center}
\tiny
\caption{Results of different ML models on multiple molecular classification benchmarks with and without molecular data augmentations. Both the average and standard deviation of the test ROC-AUC scores are reported. }
\label{tb:mol_classification}
\begin{tabular}{@{}l|lllllll@{}}
\toprule
Dataset & BBBP & BACE & ClinTox & HIV & MUV & Tox21 & SIDER \\
\# of data points & 2039 & 1513 & 1478 & 41127 & 93087 & 7831 & 1427 \\
\# of tasks & 1 & 1 & 2 & 1 & 17 & 12 & 27 \\
\cline{1-8}
Models & \multicolumn{7}{c}{ROC-AUC on test set (scaffold split)} \\
\midrule
RF & 53.34 (1.19) & 71.11 (0.46) & 50.00 (0.00) & 54.59 (1.43) & 50.00 (0.00) & 53.19 (0.49) & 52.13 (1.27) \\
RF\textsubscript{aug} & 54.29 (0.45) & 74.87 (1.83) & 50.83	(1.18) & 55.48 (0.19) & 50.62 (0.00) & 54.68 (0.78) & 53.80 (1.42) \\
SVM & 58.12 (0.00) & 73.55 (0.00) & 60.35 (0.00) & 58.75 (0.00) & 49.98 (0.00) & 55.78 (0.00) & 56.00 (0.00)  \\
SVM\textsubscript{aug} & 58.62 (0.00) & 76.70 (0.00) & 66.49 (0.00) & 62.73 (0.00) & 53.74 (0.00) & 57.83 (0.00) & 56.52 (0.00)\\
GCN & 71.82 (0.94) & 75.63 (1.95)& 67.42 (7.62) & 74.05 (3.03)&71.64 (4.00) & 70.86 (2.58)& 53.60 (3.21) \\
GCN\textsubscript{aug} & 73.66 (1.13) & 82.32 (1.87)& 75.20 (2.60)& 75.76 (0.98)& 84.68 (3.63)&74.63 (1.47)&68.99 (2.42)  \\
GINE & 71.69 (1.08) & 53.66 (5.98) & 61.40 (1.22) & 68.48 (1.51) &67.99 (5.70) & 70.24 (1.72) & 58.84 (4.19)\\
GINE\textsubscript{aug} & 73.64 (1.17) & 80.27 (2.26) & 76.81 (6.29) & 77.31 (0.32) & 80.01 (3.54) & 75.15 (1.29) & 67.97 (2.01) \\
DeeperGCN & 61.01 (3.92) & 75.73 (5.29) & 57.56 (1.40) & 65.67 (1.71) & 67.99 (5.70) & 70.78 (1.54) & 53.60 (3.34)\\
DeeperGCN\textsubscript{aug} & 70.00 (0.51) & 86.51 (0.39) & 64.23 (3.06) & 75.00 (0.75) & 81.01 (5.30) & 74.80 (1.13) & 65.45 (1.69)\\
AttentiveFP & 68.56 (1.16) & 81.39 (1.00) & 69.97 (4.53) & 75.46 (2.32) & 63.60 (7.83) & 70.04 (2.63) & 57.57 (4.13)\\
AttentiveFP\textsubscript{aug} & 73.58 (0.65) & 83.32 (0.53) & 76.07 (5.98) & 77.96 (1.27) & 78.22 (3.06) & 75.89 (1.31) & 67.90 (2.16) \\
\botrule
\end{tabular}
\end{center}
\end{table}

\begin{table}[h]
\setlength{\tabcolsep}{3pt}
\begin{center}
\footnotesize
\caption{Results of different ML models on multiple molecular regression benchmarks with and without molecular graph data augmentations. Both the average and standard deviation of the RMSEs are reported.}
\label{tb:mol_regression}
\begin{tabular}{@{}l|lllll@{}}
\toprule
Dataset & FreeSolv & ESOL & Lipo & QM7 & QM8 \\
\# of data points & 642 & 1128 & 4200 & 6830 & 21786 \\
\# of tasks & 1 & 1 & 1 & 1 & 12 \\
\cline{1-6}
Models & \multicolumn{5}{c}{RMSE on test set (scaffold split)} \\
\midrule
RF & 4.049 (0.240) & 1.591 (0.020) & 0.9613 (0.0088) & 166.8 (0.4) & 0.03677 (0.00040) \\
RF\textsubscript{aug} & 4.140 (0.007) & 1.528 (0.021) & 0.9950 (0.0041) & 168.4 (1.2) & 0.03741 (0.00011) \\
SVM & 3.143 (0.000) & 1.496 (0.000) & 0.8186 (0.0000) & 156.9 (0.0) & 0.05445 (0.00000) \\
SVM\textsubscript{aug} & 3.092 (0.000) & 1.433 (0.000) & 0.8148 (0.0000) & 169.1 (0.0) & 0.05352 (0.00000) \\
GCN & 2.847 (0.685) & 1.433 (0.067) & 0.8423 (0.0569) & 123.0 (0.9)& 0.03660 (0.00112)\\
GCN\textsubscript{aug} & 2.311 (0.198) & 1.358 (0.039) & 0.8069 (0.0104) & 120.8 (1.8) & 0.03480 (0.00081)\\
GIN & 2.760 (0.180) & 1.450 (0.021) & 0.8500 (0.0722)& 124.8 (0.7)& 0.03708 (0.00092)\\
GIN\textsubscript{aug} & 2.434 (0.051) & 1.325 (0.028) & 0.7914 (0.0310) &118.6 (1.1) & 0.03430 (0.00066)\\
DeeperGCN & 3.107 (0.462) & 1.433 (0.030) & 1.0008 (0.0150) &128.1 (2.9) &0.04224 (0.00115) \\
DeeperGCN\textsubscript{aug} & 2.368 (0.106) & 1.431 (0.057)& 0.8095 (0.0102)& 124.9 (1.1)& 0.03969 (0.00256) \\
AttentiveFP & 2.292 (0.424) & 0.884 (0.070) & 0.7682 (0.0429) & 119.4 (0.6) & 0.03466 (0.0006) \\
AttentiveFP\textsubscript{aug} & 1.779 (0.081) & 0.857 (0.016) & 0.7518 (0.0092) & 117.8 (1.6) & 0.03419 (0.0008) \\
\botrule
\end{tabular}
\end{center}
\end{table}

Besides FP augmentation, we introduce three molecular graph augmentation techniques, node masking, bond deletion, and substructure removal, focusing on GNN models \cite{you2020graphcontrstive,wang2021molclr}. 
To demonstrate how molecular graph augmentation benefits the molecular property prediction, we implement multiple GNN models, including GCN \cite{kipf2017semi}, GIN \cite{xu2018how,Hu2020Strategies}, DeeperGCN \cite{li2019deepgcns, li2020deepergcn}, and AttentiveFP \cite{xiong2019pushing}, and compare the results trained with and without augmentations. For more details of the GNN implementation and training, please refer to Supplementary Information (Section - G). Table~\ref{tb:mol_classification} shows the test ROC-AUC of GNNs model on multiple classification benchmarks. It is demonstrated that on all the 7 classification benchmarks containing 61 tasks, GNN models trained with augmented data surpasses baseline results by significant amounts. For example, GIN models trained with augmentations achieve an averaged 11.27\% improvement comparing to GIN with no augmentation. Also, molecular graph augmentation improves the test RMSE results on all the regression benchmarks as shown in Table~\ref{tb:mol_regression}. On FreeSolv data set, a small yet challenging benchmark concerning molecular hydrogen free energy in water, augmentations reduce RMSE by 0.575 in average of the 4 GNN models, which is a 20.55\% improvement with respect to the non-augmented baseline models. Such augmentation techniques not only benefit training with limited data, but also improves performance on large data sets with tens of thousands molecules, like HIV, MUV, and QM8. Results shown in Table~\ref{tb:mol_classification} and \ref{tb:mol_regression} demonstrate that our molecular graph augmentations improve the performance of challenging classification and regression property prediction tasks on various GNN models, regardless of different message passing and aggregation functions. 

\subsection{Combination of Different Graph Augmentations}
\begin{figure}[h]
\centering
      \centering
      \includegraphics[width=\linewidth]{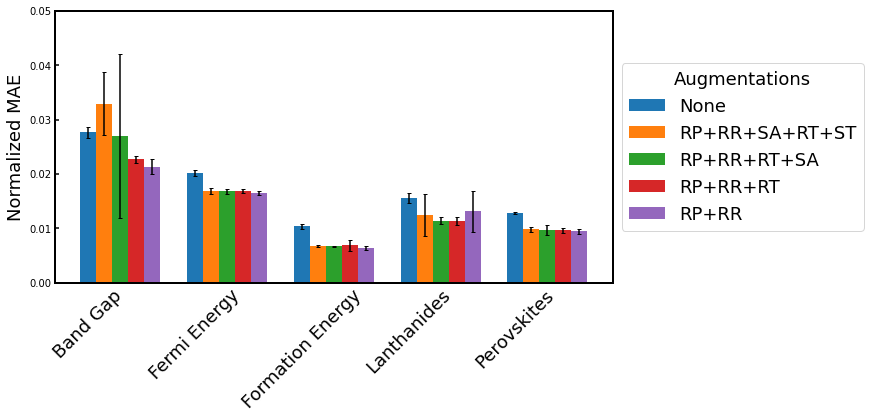}
\caption{Comparison of molecular graph augmentations and their combinations. (a) Test ROC-AUC on classification benchmarks. (b) Test normalized RMSE on regression benchmarks. }
\label{fig:ablation_crystals}
\end{figure}


\begin{figure}[h]
\centering
    \begin{subfigure}{0.87\textwidth}
      \centering
      \includegraphics[width=\linewidth]{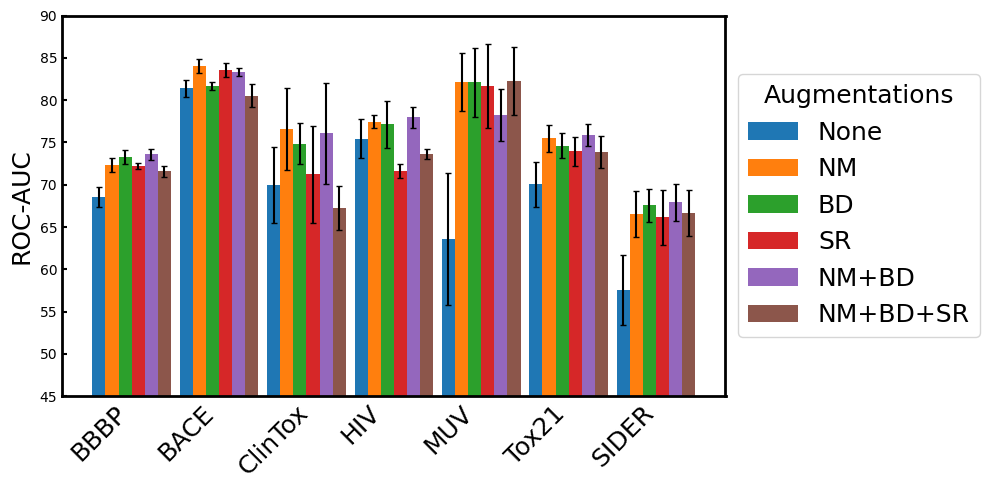}
      \caption{}
    \end{subfigure}
    \begin{subfigure}{0.87\textwidth}
      \centering
      \includegraphics[width=\linewidth]{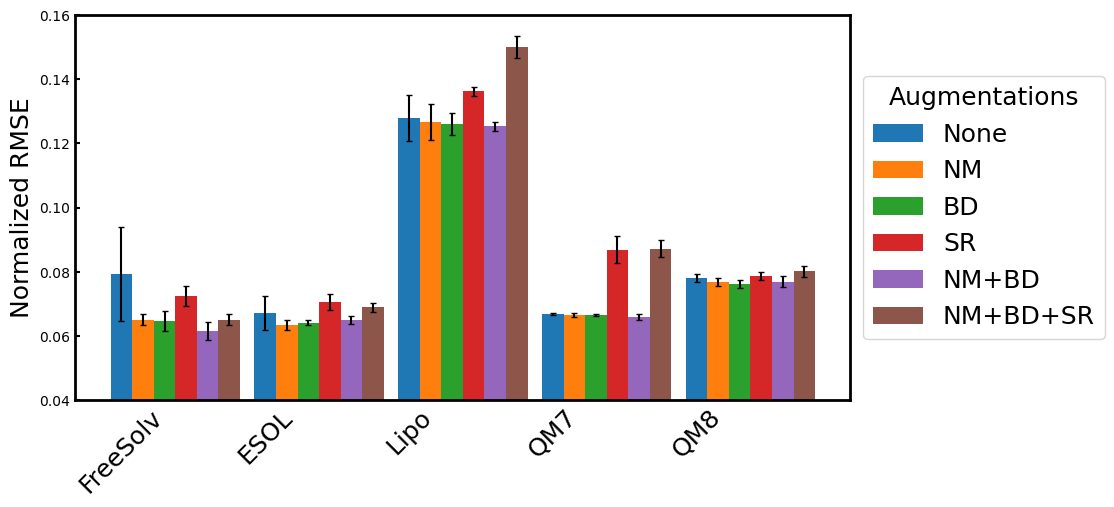}
      \caption{}
    \end{subfigure}
\caption{Comparison of Crystalline system augmentations and their combinations. The Bar plot represents the normalized MAE of the five different datasets. }
\label{fig:ablation}
\end{figure}
To understand the effect of augmentations and evaluate the best strategy for using these augmentations, we conduct a systematic ablation study for both crystals and molecules. For crystalline systems, we trained using all the five augmentation and evaluate the performance of CGCNN model using MAE as a performance metric. The plot for the normalized MAE for all the 5 datasets with different augmentation strategies has been shown in Figure~\ref{fig:ablation_crystals}. To find the optimal augmentation strategy we experiment with different augmentation strategies: no augmentation (None), random perturbation/random rotation(RP+RR), random perturbation/ random rotation/random translation(RP+RR+RT),  random perturbation/ random rotation/random translation/swap axes (RP+RR+RT+SA), all augmentations(RP+RR+RT+SA+ST). To conduct the ablation all the training parameters were kept the same to ensure fair comparison among the augmentation strategies. Finally, we select the random perturbation, random rotation and swap axes along with original data as our final augmentation strategy to ensure  computational efficiency and higher performance when training our GNNs and fingerprint based ML models. However, we would like to note that an optimal  augmentation strategy is task specific and data dependent.

To investigate the optimal augmentation strategy in molecules, we compare the performance of different molecular graph augmentations as well as their combinations. Figure~\ref{fig:ablation} shows the test results, where mean and standard deviation of 3 individual runs are illustrated by the height of each column and error bar respectively. In particular, 6 augmentation strategies are considered: no augmentation (None), node masking (NM), bond deletion (BD), substructure removal (SR), node masking with bond deletion (NM+BD), and all augmentations combined (NM+BD+SR). For comparison, AttentiveFP models with the same settings are implemented for each benchmark with different augmentations. More implementation details and results concerning augmentation combinations can be found in Supplementary Information (Table S4 and Table S5). 

Test ROC-AUC of classification tasks are illustrated in Figure~\ref{fig:ablation}(a). Augmentations, regardless of the specific strategies, improve the performance on most benchmarks. When applying one augmentation solely, NM and BD are generally better choices comparing to the SR. Further, the combination of NM and BD shows rival improvement as NM or BD, and even surpasses the single augmentation strategy in multiple benchmarks, including BBBP, HIV, Tox21, and SIDER. Figure~\ref{fig:ablation}(b) demonstrates the test normalized RMSE of different augmentations on regression tasks. Normalized RMSE is calculated by dividing the RMSE by the range of label in each dataset. BD achieves the best performance among NM, BD, and SR in most benchmarks. Results from NM are also close to BD and even better in Lipo dataset. Similarly, NM with BD obtains competitive performance and surpasses other strategies in 3 out of all 5 benchmarks. In general, NM, BD, and the combination of these two are the best performing augmentation strategies in most classification and regression benchmarks. However, it should be pointed out that the optimal augmentation strategy is task-dependent. For instance, though SR struggles to compete with other augmentations in most cases, in BACE and MUV, SR still achieves comparable improvement to NM and BD. Also, combination of all 3 molecular graph augmentations does not perform well compared to other strategies. This maybe due to such combinations generating augmented graphs that are hard to match with the original graph.

\subsection{AugLiChem Package}\label{subsec3}


The package is designed with ease of use in mind for all aspects.
Publicly available data is downloaded and cleaned automatically, significantly reducing the time to start running experiments.
Implementations of popular models are available for easy comparison. Both random splitting and k-fold cross validation can be done for crystal data sets. Single and multi-target training can be done with molecule data sets, where handling missing data is done automatically. These training styles are included for ease of experimentation. Both molecule and crystal data handling is also designed so only the training set is augmented, ensuring evaluation validity. Examples can be found in Supplementary Information (Section - H). Detailed usage guides and documentation for all relevant functions and models: \url{https://baratilab.github.io/AugLiChem/}.

\section{Conclusion}\label{sec6}

The work introduces a systematic methodology for augmentation of molecular data sets. We design augmentation strategies for crystalline systems and molecules. Using these augmentation, we are able to significantly boost the amount of data available for the machine learning models. For crystalline systems, we design five different augmentation strategies for crystalline systems and select the most effective ones to train ML models. Using the augmented data we observe significant gains in performance of four different graph neural networks on five different data sets consisting a wide range of properties of crystalline systems. Similarly, for molecules we investigate two augmentation strategies for fingerprint-based ML and three strategies for GNN models. We observe performance gains on various popular benchmarks (seven classification and five regression data sets in total) when compared to performance of baseline GNNs without augmentation, indicating the effectiveness of augmentation strategies. The performance gains observed when training GNN models empirically proves the effectiveness of data augmentation strategies on crystalline and molecular data sets. Finally, we also develop an open source python package that implements all the augmentations introduced in the paper and can be imported directly into any machine learning workflow. 


\section{Data Availability Statement}
The data that support the findings of this study are openly available in Materials Project (reference number: 79), Molecule Net Benchmark(reference number: 4), Perovskites dataset (reference number: 80), Lanthanides dataset (reference number: 8).
\backmatter











\bibliography{sn-bibliography}


\end{document}



\title[AugLiChem SI]{AugLiChem: Data Augmentation Library of Chemical Structures for Machine Learning: Supplementary Information}


\author[1]{\fnm{Rishikesh} \sur{Magar}}
\equalcont{These authors contributed equally to this work.}

\author[1]{\fnm{Yuyang} \sur{Wang}}
\equalcont{These authors contributed equally to this work.}

\author[1]{\fnm{Cooper} \sur{Lorsung}}
\equalcont{These authors contributed equally to this work.}

\author[2]{\fnm{Chen} \sur{Liang}}

\author[1]{\fnm{Hariharan} \sur{Ramasubramanian}}

\author[2]{\fnm{Peiyuan} \sur{Li}}

\author*[1,2]{\fnm{Amir} \sur{Barati Farimani}}\email{barati@cmu.edu}

\affil[1]{\orgdiv{Department of Mechanical Engineering}, \orgname{Carnegie Mellon University}, \orgaddress{ \city{Pittsburgh}, \state{PA}, \country{USA},\postcode{15213}}}

\affil[2]{\orgdiv{Department of Chemical Engineering}, \orgname{Carnegie Mellon University}, \orgaddress{ \city{Pittsburgh}, \state{PA}, \country{USA},\postcode{15213}}}

\maketitle

\newpage

\section{Detailed Results of Molecular FP Augmentations}

Table~\ref{tb:fP_cls} and Table~\ref{tb:fP_rgr} show the detailed results of molecular FP augmentation on classification and regression benchmarks, respectively. The performance of both FP Break and FP Concat is reported. 

\begin{table}[h]
\setlength{\tabcolsep}{3pt}
\begin{center}
\tiny
\caption{Results of different GNN models on multiple molecular classification benchmarks with and without molecular graph data augmentations. Both the average and standard deviation of the ROC scores are reported. }
\label{tb:fP_cls}
\begin{tabular}{@{}l|lllllll@{}}
\toprule
Dataset & BBBP & BACE & ClinTox & HIV & MUV & Tox21 & SIDER \\
\# of molecules & 2039 & 1513 & 1478 & 41127 & 93087 & 7831 & 1427 \\
\# of tasks & 1 & 1 & 2 & 1 & 17 & 12 & 27 \\
\cline{1-8}
Models & \multicolumn{7}{c}{Prediction on test set (scaffold split)} \\
\midrule
RF & 53.34 (1.19) & 71.11 (0.46) & 50.00 (0.00) & 54.59 (1.43) & 50.00 (0.00) & 53.19 (0.49) & 52.13 (1.27) \\
RF\textsubscript{FP\_Break} & 53.84 (0.28) & 73.94 (1.01) & 50.83 (1.18) & 55.22 (0.72) & 50.00 (0.00) & 53.71 (0.70) & 53.80 (1.42) \\
RF\textsubscript{FP\_Concat} & 54.29 (0.45) & 74.87 (1.83) & 50.00 (0.00) & 55.48 (0.19) & 50.62 (0.00) & 54.68 (0.78) & 53.52 (1.78) \\
SVM & 58.12 (0.00) & 73.55 (0.00) & 60.35 (0.00) & 58.75 (0.00) & 49.98 (0.00) & 55.78 (0.00) & 56.00 (0.00)  \\
SVM\textsubscript{FP\_Break} & 58.12 (0.00) & 76.70 (0.00) & 54.88 (0.00) & 59.38 (0.00) & 50.61 (0.00) & 56.12 (0.00) & 56.52 (0.00)\\
SVM\textsubscript{FP\_Concat} & 56.62 (0.00) & 76.27 (0.00) & 66.49 (0.00) & 62.73 (0.00) & 53.74 (0.00) & 57.83 (0.00) & 55.18 (0.00)\\
\botrule
\end{tabular}
\end{center}
\end{table}

\begin{table}[h]
\begin{center}
\footnotesize
\caption{Results of different GNN models on multiple molecular regression benchmarks with and without molecular graph data augmentations. Both the average and standard deviation of the RMSEs are reported.}
\label{tb:fP_rgr}
\begin{tabular}{@{}l|lllll@{}}
\toprule
Dataset & FreeSolv & ESOL & Lipo & QM7 & QM8 \\
\# of molecules & 642 & 1128 & 4200 & 6830 & 21786 \\
\# of tasks & 1 & 1 & 1 & 1 & 12 \\
\cline{1-6}
Models & \multicolumn{5}{c}{Prediction on test set (scaffold split)} \\
\midrule
RF & 4.049 (0.240) & 1.591 (0.020) & 0.9613 (0.0088) & 166.8 (0.4) & 0.03677 (0.00040) \\
RF\textsubscript{FP\_Break} & 4.159 (0.011) & 1.528 (0.021) & 1.0094 (0.0042) & 168.4 (1.2) & 0.03741 (0.00011) \\
RF\textsubscript{FP\_Concat} & 4.140 (0.007) & 1.644 (0.015) & 0.995 (0.004) & 169.5 (0.5) & 0.03785 (0.00054) \\
SVM & 3.143 (0.000) & 1.496 (0.000) & 0.8186 (0.0000) & 156.9 (0.0) & 0.05445 (0.00000) \\
SVM\textsubscript{FP\_Break} & 3.092 (0.000) & 1.433 (0.000) & 0.8591 (0.0000) & 169.1 (0.0) & 0.05352 (0.00000) \\
SVM\textsubscript{FP\_Concat} & 3.131 (0.000) & 1.491 (0.000) & 0.815 (0.000) & 169.7 (0.0) & 0.06150 (0.00000) \\
\botrule
\end{tabular}
\end{center}
\end{table}

\newpage
\section{Ablation study for Crystalline Systems augmentations}
To study the effect of the augmentation that we have developed for crystalline systems, we conducted a systematic ablation study. For all the GNN models, we initially use all the augmentation strategies and evaluate the MAE. In general, using all the augmentation strategies has the highest MAE among our experiments. Subsequently, we remove one augmentation for every experiment and identify the most effective ones. It was observed that the augmentation strategies SuperCell Transform and Translation transformation do not perform particularly well. Therefore, when performing our augmentation experiments we use only perturbation, rotation and swap axes transform 

\begin{table}[h]
\setlength{\tabcolsep}{3pt}
\begin{center}
\footnotesize
\caption{Results of combinations of different molecular data augmentations: node masking (NM), bond deletion (BD), and substructure removal (SR), on multiple molecular regression benchmarks. Both the average and standard deviation of the RMSEs are reported. }
\label{tb:ablation_rgr}
\begin{tabular}{@{}lllll|lllll@{}}
\toprule
\multicolumn{5}{l|}{Dataset} & Band Gap & Fermi Energy & Formation energy & Lanthanides & Perovskites \\
\multicolumn{5}{l|}{\# of molecules} & 26709 & 27779 & 26078 & 4133 & 18928 \\
\multicolumn{5}{l|}{\# of tasks} & 1 & 1 & 1 & 1 & 1 \\
\cline{1-10}
P & R & Su & Sw & T & \multicolumn{5}{c} {Prediction on test set (Random Split)} \\
\midrule
\cmark & \cmark & \cmark & \cmark & \cmark & 0.5460(0.257)  & 0.4216(0.005) &0.0368(0.0004)  & 0.0882(0.012) & 0.0569(0.0009) \\
\cmark & \cmark & \xmark & \cmark & \cmark & 0.4477(0.098) & 0.4203(0.004) & 0.0365(0.0001) & 0.0805(0.001) & 0.0567(0.002)\\
\cmark & \cmark & \xmark & \xmark & \cmark & 0.3762(0.004) &0.4235(0.004)  & 0.0374 (0.002)& 0.0805(0.003) & 0.0557(0.001) \\
\cmark & \cmark & \xmark & \xmark & \xmark & 0.3546(0.009) &0.4129(0.003)& 0.0350(0.0007) & 0.0927(0.010) & 0.0548(0.0009)\\
\botrule
\end{tabular}
\end{center}
\end{table}

\newpage
\section{Ablation Study of Molecular Graph Augmentations}

In this section, we report the raw results concerning the ablation study of molecular graph augmentations in corresponding to Figure 4 in the main manuscript. Table~\ref{tb:ablation_cls} shows the results of ablation study for classification. By combinations of three augmentations, NM, BD and SR, we build five different augmentation models in total: NM only, BD only, SR only, NM+BD, NM+BD+SR. These models are implemented for 7 classification benchmarks from MoleculeNet and we use AttentiveFP model. Table~\ref{tb:ablation_rgr} shows the results of ablation study for regression. The models are implemented for 5 regression benchmarks from MoleculeNet. Also, the first line is results for supervised learning trained with initial dataset and the next five lines are for different combinations of augmentation. By these results, we can come up with that the performance for different combinations change from different datasets.

\begin{table}[h]
\setlength{\tabcolsep}{3pt}
\begin{center}
\tiny
\caption{Results of combinations of different molecular data augmentations: node masking (NM), bond deletion (BD), and substructure removal (SR), on multiple molecular classification benchmarks. Both the average and standard deviation of the ROC scores are reported. }
\label{tb:ablation_cls}
\begin{tabular}{@{}lll|lllllll@{}}
\toprule
\multicolumn{3}{l|}{Dataset} & BBBP & BACE & ClinTox & HIV & MUV & Tox21 & SIDER \\
\multicolumn{3}{l|}{\# of molecules} & 2039 & 1513 & 1478 & 41127 & 93087 & 7831 & 1427 \\
\multicolumn{3}{l|}{\# of tasks} & 1 & 1 & 2 & 1 & 17 & 12 & 27 \\
\cline{1-10}
NM & BD & SR & \multicolumn{7}{c}{Prediction on test set (scaffold split)} \\
\midrule
\xmark & \xmark & \xmark & 68.56 (1.16) & 81.39 (1.00) & 69.97 (4.53) & 75.46 (2.32) & 63.60 (7.83) & 70.04 (2.63) & 57.57 (4.13)\\
\cmark & \xmark & \xmark &72.33 (0.84)&84.04 (0.85)&76.62 (4.84)&77.48 (0.73)&82.12 (3.39)&75.47 (1.58)&66.54 (2.72) \\
\xmark & \cmark & \xmark & 73.31 (0.82)&81.68 (0.50)&74.86 (2.46)&77.13 (2.81)&82.09 (4.07)&74.62 (1.44)&67.56 (1.94)\\
\xmark & \xmark & \cmark &72.24 (0.33)&83.54 (0.80)&71.23 (5.72)&71.62 (0.86)&81.70 (4.95)&73.93  (1.66)&66.14 (3.26) \\
\cmark & \cmark & \xmark & 73.58 (0.65) & 83.32 (0.53) & 76.07 (5.98) & 77.96 (1.27) & 78.22 (3.06) & 75.89 (1.31) & 67.90 (2.16)\\
\cmark & \cmark & \cmark & 71.57 (0.65)&80.53 (1.32)& 67.25 (2.55)&73.59 (0.61)&82.26 (4.04)&73.87 (1.88)&66.70 (2.71)\\
\botrule
\end{tabular}
\end{center}
\end{table}

\begin{table}[h]
\setlength{\tabcolsep}{3pt}
\begin{center}
\footnotesize
\caption{Results of combinations of different molecular data augmentations: node masking (NM), bond deletion (BD), and substructure removal (SR), on multiple molecular regression benchmarks. Both the average and standard deviation of the RMSEs are reported. }
\label{tb:ablation_rgr}
\begin{tabular}{@{}lll|lllll@{}}
\toprule
\multicolumn{3}{l|}{Dataset} & FreeSolv & ESOL & Lipo & QM7 & QM8 \\
\multicolumn{3}{l|}{\# of molecules} & 642 & 1128 & 4200 & 6830 & 21786 \\
\multicolumn{3}{l|}{\# of tasks} & 1 & 1 & 1 & 1 & 12 \\
\cline{1-8}
NM & BD & SR & \multicolumn{5}{c}{Prediction on test set (scaffold split)} \\
\midrule
\xmark & \xmark & \xmark & 2.292 (0.424) & 0.884 (0.070) & 0.7682 (0.0429) & 119.4 (0.6) & 0.03466 (0.0006)\\
\cmark & \xmark & \xmark &1.882 (0.046)&0.835 (0.022)&0.7601 (0.0331)&119.0 (1.0)&0.03412 (0.0005) \\
\xmark & \cmark & \xmark & 1.869 (0.091)&0.845 (0.011)& 0.7563 (0.0200)&118.9 (0.6)&0.03380 (0.0006)\\
\xmark & \xmark & \cmark & 2.095 (0.092)& 0.929 (0.033)&0.8180 (0.0085)&155.2 (7.5)& 0.03491 (0.0006)\\
\cmark & \cmark & \xmark & 1.779 (0.081) & 0.857 (0.016) & 0.7518 (0.0092) & 117.8 (1.6)& 0.03419 (0.0008) \\
\cmark & \cmark & \cmark &1.882 (0.047)&0.909 (0.018)&0.9011 (0.0204)&155.9 (4.7)& 0.03560 (0.0007) \\
\botrule
\end{tabular}
\end{center}
\end{table}


\newpage

\section{Details of Benchmarks - Crystalline}

\begin{table}[htb!]
  \centering
  \begin{tabular}{llllll}
    \toprule
    Dataset & \# Molecules & \# Tasks & Task type & Metric & Split \\
    \midrule
    Band Gap & 26709 & 1 & Regression & MAE & Random \\
    Fermi Energy & 27779 & 1 & Regression & MAE & Random \\
    Formation Energy & 26078 & 1 & Regression & MAE & Random \\
    Lanthanides & 4133 & 1 & Regression & MAE & Random \\
    Perovskites & 18928 & 1 & Regression & MAE & Random \\
    \bottomrule
  \end{tabular}
  \caption{Overview of all crystalline benchmarks used in this work.}
  \label{tb:benchmarks}
\end{table}

\newpage
\section{Details of Benchmarks - Molecule}

\begin{table}[htb!]
  \centering
  \begin{tabular}{llllll}
    \toprule
    Dataset & \# Molecules & \# Tasks & Task type & Metric & Split \\
    \midrule
    SIDER & 1427 & 27 & Classification & ROC-AUC & Scaffold \\
    ClinTox & 1478 & 2 & Classification & ROC-AUC & Scaffold \\
    BACE & 1513 & 1 & Classification & ROC-AUC & Scaffold \\
    BBBP & 2039 & 1 & Classification & ROC-AUC & Scaffold \\
    Tox21 & 7831 & 12 & Classification & ROC-AUC & Scaffold \\
    HIV & 41127 & 1 & Classification & ROC-AUC & Scaffold \\
    MUV & 93087 & 17 & Classification & ROC-AUC & Scaffold \\
    \midrule
    FreeSolv & 642 & 1 & Regression & RMSE & Scaffold \\
    ESOL & 1128 & 1 & Regression & RMSE & Scaffold \\
    Lipo & 4200 & 1 & Regression & RMSE & Scaffold \\
    QM7 & 6830 & 1 & Regression & RMSE & Scaffold \\
    QM8 & 21786 & 12 & Regression & RMSE & Scaffold \\
    \bottomrule
  \end{tabular}
  \caption{Overview of all molecular benchmarks used in this work.}
  \label{tb:benchmarks}
\end{table}

\newpage
\section{Training Details - Crystalline Systems}
We used a set of different hyperparameters depending on type of GNN model used for training all the 5 datasets. For the CGCNN model\cite{xie2018cgcnn} we used the tuned hyperparmeters in the original work except we trained the model for 50 epochs and used Adam optimizer for stability\cite{kingma2014adam}. It must be noted that baseline CGCNN model without taking the augmented data for training was also using the same hyperparameter setting. For MEGNET model the default hyperparmeters were used as reported in the orginal work with number of training epochs for the model to be set to 50\cite{chen2019graph}. The SchNet model and GIN model were developed using pytorch geometric\cite{fey2019fast_pyG}. We use 5 GIN layers with an embedding dimension of 128. We use the Adam optimizer to train the model with an initial learning rate of 0.001 and weight decay $10^{-6}$. To train the SchNet model, we use a 128 hidden channels, 128 filters as parameters. A maximum of 12 neighbors is considered in a cut off radius of $8\r{A}$, the readout function is summation over all the nodes. The Schnet model was trained with Adam optimizer\cite{kingma2014adam} or 100 epochs with an initial learning rate of $0.001$ and weight decay $10^{-6}$

The protocol for training the shallow machine learning models was exactly same as the molecules except we are using AGNI fingerprints that can effectively capture crystalline systems \cite{botu2017machine}

\newpage
\section{Training Details - Molecule}

On each task in molecular benchmarks, we train the GNN model via Adam optimizer \cite{kingma2014adam} with initial learning rate $5 \times 10^{-4}$ and weight decay $10^{-5}$ for 100 epochs. All GNN models, namely GCN \cite{kipf2017semi}, GIN \cite{xu2018how, Hu2020Strategies}, DeepGCN \cite{li2019deepgcns}, and AttentiveFP \cite{xiong2019pushing}, are set with node embedding dimension 128. Batch size is set to 64 and dropout ratio is set to 0 for regression benchmarks and 0.2 for classification benchmarks by default. GCN, GIN, and DeepGCN are developed with 3, 5, and 28 graph convolutional layers respectively, following the general setting for each model. Besides, when training AttentiveFP, we set the hyperparameters following the optimal settings reported in the original work \cite{xiong2019pushing}. It should be pointed out that, with further hyperparameter search, GNN models may perform better on certain molecular benchmarks. However, models trained under settings reported in our work have demonstrated the effectiveness of data augmentations for chemical structures, which is one of the major contributions of our work. 

For shallow ML models: random forest (RF) and support vector machine (SVM), we embed molecules into vectors through RDKFP or ECFP \cite{rogers2010extended}. We train each ML model under different settings and select the best performing model on the validation set. Results of the selected model on test set are reported. RF models are trained with number of estimators from $\{10, 50, 100\}$ and maximal depth from $\{5, 10, 15\}$. SVM models are trained with $C$ from $\{0.1, 1, 10, 100\}$ with either polynomial or radial basis function (RBF) kernel.

\newpage
\section{Package Structure}

\noindent
\begin{lstlisting}[language=Python, numbers=none caption=AugLiChem package structure]
AugLiChem/
  -.travis.yml  -codecov.yml  -LICENSE  -README.md  -setup.py
  -auglichem/
    -crystal/
      -__init__.py  -_transforms.py
      -data/
        -__init__.py  -_load_sets.py  -_crystal_dataset.py
      -models/
        -__init__.py  -cgcnn.py  -gin.py  -schnet.py
    -molecule/
      -__init__.py -_compositions.py  -_transforms.py
      -data/
        -__init__.py  -_load_sets.py  -_molecule_dataset.py
      -models/
        -__init__.py  -afp.py  -deepgcn.py  -gcn.py  -gine.py
    -test/
      -test_crystal.py  -test_molecule.py  -test_crystal.py  -test_utils.py
    -utils/
      -__init__.py  -_constants.py  -_splitting.py
\end{lstlisting}

\subsection{Usage Examples}
 We see below random atom masking with 10\% of atoms masked for the HIV dataset implemented.\\

 \noindent
 \begin{lstlisting}[language=Python, caption=Molecule Data Example]
 from auglichem.molecule.data import MoleculeDatasetWrapper
 from auglichem.molecule import RandomAtomMask

 transform = RandomAtomMask(0.1)
 dataset = MoleculeDatasetWrapper("HIV", transform=transform, batch_size=64)
 train_loader, valid_loader, test_loader = dataset.get_data_loaders()
 \end{lstlisting}

 \noindent
 Inorganic augmentations require explicitly calling the augmentation rather than being done at call-time.
 This can bee seen in the code snippet below:\\

 \noindent
 \begin{lstlisting}[language=Python, caption=Crystal Data Example]
 from auglichem.crystal.data import CrystalDatasetWrapper
 from auglichem.crystal import SupercellTransformation

 transform = SuperCellTransformation()
 dataset = CrystalDatasetWrapper("lanthanides", batch_size=1024)
 train_loader, valid_loader, test_loader = dataset.get_data_loaders(transform)
 \end{lstlisting}
 
 \subsection{Data Splitting and Augmentation}
 
 Data splitting and cleaning is handled automatically by AugLiChem.
 Cleaning is done before splitting, where invalid SMILES strings are removed for molecular data, and crystal data sets have been cleaned before download time.
 Splitting the data also sets up the augmentation.
 For molecular systems, this is simple, and the molecules are augmented at call time.
 That is, when calling the data from the `MoleculeDataset` object, wrapper, or loader, the graph representation is built, and then augmented.
 Augmentation is done this way because augmented molecules do not have valid SMILES representations, and cannot easily be stored for later use.
 Crystal data is augmented before splitting because augmented crystals can be saved next to the originals as CIF files.
After the data is augmented, augmented CIF files are added to the training set.
CIF files are loaded at call time.
Because the augmentation has already been done, only building the graph representation is done before returning the data.

\newpage

\bibliography{sn-bibliography}
